\documentclass[final]{article}

\usepackage[nonatbib]{neurips_2019}

\usepackage{booktabs}       
\usepackage{amsfonts}       
\usepackage{nicefrac}       
\usepackage{microtype}      
\usepackage{graphicx}       
\usepackage{bm}             
\usepackage{amsmath}        
\usepackage{algorithm}      
\usepackage{algorithmic}    
\usepackage{lipsum}
\usepackage{multicol}
\usepackage{enumitem}

\usepackage[backend=bibtex]{biblatex} 
\graphicspath{ {./images/} }

\title{Making Good on LSTMs' Unfulfilled Promise}

\author{
  Daniel Philps 
  \\
  Rothko Investment Strategies\\ 
  Department of Computer Science\\
  City, University of London\\
  \texttt{daniel.philps@city.ac.uk} 
  \And
  Artur d'Avila Garcez \\
 Department of Computer Science\\
  City, University of London\\
  \texttt{a.garcez@city.ac.uk} 
  \And
  Tillman Weyde \\
  Department of Computer Science\\
  City, University of London\\
  \texttt{t.e.weyde@city.ac.uk} 
}  

\begin{document}

\maketitle

\begin{abstract}
LSTMs promise much to financial time-series analysis, temporal and cross-sectional inference, but we find that they do not deliver in a real-world financial management task. We examine an alternative called Continual Learning (CL), a memory-augmented approach, which can provide transparent explanations, i.e. which memory did what and when. 
This work has implications for many financial applications including credit, time-varying fairness in decision making and more. 
We make three important new observations. 
Firstly, as well as being more explainable, time-series CL approaches outperform LSTMs as well as a simple sliding window learner using feed-forward neural networks (FFNN). 
Secondly, we show that CL based on a sliding window learner (FFNN) is more effective than CL based on a sequential learner (LSTM). Thirdly, we examine how real-world, time-series noise impacts several similarity approaches used in CL memory addressing. 
We provide these insights using an approach called Continual Learning Augmentation (CLA) tested on a complex real-world problem, 
emerging market equities investment decision making. CLA provides a test-bed as it can be based on different types of time-series learners, allowing testing of LSTM and FFNN learners side by side. 
CLA is also used to test several distance approaches used in a memory recall-gate: Euclidean distance (ED), dynamic time warping (DTW), auto-encoders (AE) and a novel hybrid approach, \emph{warp-AE}. 
We find that ED under-performs DTW and AE but \emph{warp-AE} shows the best overall performance in 
a real-world financial task.  

\textbf{Keywords}: Continual learning, time-series, LSTM, similarity, DTW, auto-encoder
\end{abstract}

\section{Introduction}

Both LSTMs \cite{Hochreiter_1997} and a wide range of time-series $sliding$ $window$ approaches suffer from a common problem
: 
how to deal with long versus short term dependencies \cite{Koutn14,Mozer:1991:IMT:2986916.2986950,Gers_Schidhuber_LSTMs_Timeseries_2001}. 
This problem is a manifestation of $catastrophic$ $forgetting$ (CF), one of the major impediments to the development of artificial general intelligence (AGI), where the ability of a learner to generalise to older tasks is corrupted by learning newer tasks \cite{French1999CatastrophicFI,MCCLOSKEY1989109}.
To address this problem, Continual Learning (CL) has been developed. Although very few time-series CL approaches exist, some have the advantage of having interpretable memory addressing \cite{Philps_2018}, in contrast to LSTMs.
The advantages of better interpretability are significant for real-world financial problems but there are a number of open questions for time-series CL.
Firstly, should a financial time-series CL approach be based on recurrent (e.g. LSTM) or a sliding window architecture (e.g. feed-forward neural net (FFNN))? 
What impact does time-series noise have on the memory functions of a CL approach? 
How would these choices translate to performance in a real-world financial problem when compared with LSTMs and a simple sliding window approach (FFNN)?
This study empirically examines these questions in the context of the complex real-world problem of  stock selection investment decisions in emerging market equities.
\par
This paper is organised as follows.
Section \ref{sec_back} introduces common design choices for time-series CL. 
Section \ref{sec_litrev} reviews related work while
section \ref{sec_CLA} discusses \emph{Continual Learning Augmentation} remember-gates, recall-gates and memory balancing. 
Section \ref{sec_Exp} describes the experimental setup, results and interpretability of the complex, real-world tests conducted, while
section \ref{sec_Conclusion} concludes this paper.

\par
\par

\section{Background} \label{sec_back}
Machine learning based time-series approaches can be broadly separated into \emph{sliding window}: dividing a time-series into a number of discrete modelling steps, e.g. FFNN; and \emph{sequential}: attempting to model a time-series process as a sequence of values, e.g. LSTM.

A sliding window allows the choice of a wide range of learners, such as OLS regression \cite{Fama93commonrisk} or feed forward neural networks (FFNN) applied in a step forwards fashion. Specialist models have also been developed, such as time delayed neural nets (TDNN) \cite{Waibel_TDNN_1989} but these are still constrained by choices relating to the time-delays to use.   
The major shortcoming of sliding-window approaches is that, as time-steps on, all information that moves out of the sliding window is typically forgotten. 

Sequential approaches, while still requiring a sliding window for longer series, attempt to avoid the arbitrary, window sizing problem. 
However, this comes with a greater degree of complexity \cite{Zhang_RNNComplexity_Skip_16}, as cross-sectional, temporal and short versus long-term dependencies need to be established sequentially. While there are many types of sequential learner \cite{Thomas_Sequential_2002}, LSTMs are probably the most popular. 
While they are able to solve time-series problems that sliding window approaches cannot, sliding window approaches have outperformed LSTMs on seemingly more simple time-series problems \cite{Gers_Schidhuber_LSTMs_Timeseries_2001}. 
Interpreting LSTMs is also challenging \cite{Guo2019_}.
In either case, FFNNs and LSTMs applied to a time-evolving data-set, are exposed to CF \cite{Schak_2019}; occurring when a learner is applied in a $step forwards$ fashion. 
For example, a FFNN that is first trained to accurately approximate a model in a time-period $A$, and is then trained in time-period $B$, may see a deterioration in accuracy when applied to time-period $A$ again. 
CL has been proposed to address CF, using implicit or explicit memory for the purpose. 
In many cases $similarity$ plays a part in driving this \cite{liu_2015,Fei:2016:LCB:2939672.2939835,Shu_2018} but in the real-world of complex, noisy time-series similarity can be difficult to define. 
Temporal dependencies, changing modalities and 
other issues 
\cite{Schlimmer1986} all complicate gauging time-varying similarity and generally add computational expense. 
The impact of these temporal dynamics on a learner is sometimes called \emph{concept drift} \cite{Schlimmer1986,Widmer1996}.





This paper addresses whether a sequential or sliding window approach is best in a noisy real-world financial task and examines time-series similarity in a CL context. We test a sliding window approach (FFNN) and then a sequential learner (LSTM) finding that the FFNN applied as a sliding window is the best performer. Secondly, we test different time-series similarity approaches, which are used to drive CLA's memory recall-gate. We find simple Euclidean distance (ED) under-performs noise invariant similarity approaches; dynamic time warping (DTW) and auto-encoders (AE). We find the best performing similarity approach is a novel hybrid; \emph{warp-AE}.

\section{Related Work} \label{sec_litrev}
While CF remains an open problem, many techniques have been developed to address it including gated neural networks \cite{Hochreiter_1997}, explicit memory structures \cite{Weston}, prototypical addressing \cite{Snell_2017}, weight adaptation \cite{Hinton_Distilling_2015,Sprechmann_2018}, task rehearsal \cite{Silver_2002} and encoder based lifelong learning \cite{Triki2017} to name a few. 
As researchers have addressed the initial challenges of CL, other problems have emerged, such as the overhead of external memory structures \cite{Rae_2016_sparsereads}, problems with weight saturation \cite{Kirkpatrick_2017}, transfer learning \cite{Lopez-Paz_2017} and the drawbacks of outright complexity \cite{DBLP:journals/corr/ZarembaS15}. 
While most CL approaches aim to learn sequentially, only a fraction of CL approaches have been focused on time-series \cite{Kadous_TS_2002:,Graves:2006:CTC:1143844.1143891,Lipton_TS_Modeling,Thomas_TS_2017}. 
It is also unclear how effective these approaches would be when applied to open-world, state-based, temporal learning in long term, noisy, non-stationary time-series, particularly those commonly found in finance. 
As most CL approaches have been developed for well-defined, generally labelled and typically stylized tasks \cite{Lopez-Paz_2017} a strong motivation exists to develop time-series specific CL. 

\subsection{Remembering}
Regime switching models \cite{KimNelson1999} and change point detection \cite{Pettitt1979} provide a simplified answer to identifying changing states in time-series with the major disadvantage that change points between regimes (or states) are notoriously difficult to identify out of sample  \cite{fabozzi2010quantitative} and existing econometric approaches are limited by long term, parametric assumptions in their attempts \cite{Engle_1999,Zhang_2010,Siegmund_2013}. 
There is also no guarantee that a change point represents a significant change in the accuracy of an applied model, a more useful perspective for learning different states. 

\emph{Residual change} aims to observe change in the absolute error of a learner, aiming to capture as much information as possible regarding changes in the relation between independent and dependent variables. 
Different forms of residual change have been developed  \cite{Brown_1975,Jandhyala_1986,Jandhyala_1989,MacNeilt_1985,Bai_1991,Gama_COnceptDriftAdapt_2014}.   
However, most approaches assume a single or known number of change points in a series and are less applicable to a priori change points or multivariate series \cite{Yu_2007}.
\emph{Drift adaptation} approaches attempt to address some of these issues \cite{Gama_COnceptDriftAdapt_2014} but are of limited value to CL as they are focused on classification \cite{8571222}, tend to use simple and generally instance-based memory \cite{Widmer1996,Maloof2000,Klinkenberg_2004,Gomes_2010},  while tending to neglect recurring concepts \cite{8571222} and to suffer from \emph{gradual forgetting} \cite{Koychev00gradualforgetting}. 
This contrasts with explicit, task-oriented, memory structures of the sort used by CL to address CF. 
With the advent of time-series CL, residual change has another interesting application.


\subsection{Recalling} 
CL approaches that use external memory structures require an appropriate addressing mechanism 
to store and recall memories. 
Memory addressing is generally based on a similarity measure such as cosine similarity \cite{Graves_14,graves2016hybrid,Park_2017} kernel weighting \cite{Vinyals_2016}, use of linear models \cite{Snell_2017} or instance-based similarities, many using K-nearest neighbours \cite{Kaiser_2017,Sprechmann_2018}. 
More recently, auto-encoders (AE) have been used to gauge similarity in the context of multitask learning (MTL) \cite{Aljundi17} and for memory consolidation \cite{Triki2017}.   
However, these methods, as they have been applied, are not obviously well suited to assessing similarity in noisy multivariate time-series. 
In contrast, \emph{data mining} researchers have extensively researched noise invariant time-series distance measures \cite{Cha2007}, generally for time-series classification (TSC). 
While simple Euclidean distance (ED) offers a rudimentary approach for comparing time-series it has a high sensitivity to the timing of data-points, something that has been addressed by dynamic time warping (DTW) \cite{Sakoe1978}. 
However, DTW requires normalized data and is computationally expensive, although some mitigating measures have been developed \cite{Zhang:2017:DTW:3062405.3062585}. 
A relatively small subset of data-mining research has used deep learning based approaches, such as convolutional neural nets (CNN) \cite{Zheng_CNN_2014}. 
While results have been encouraging, interpretability is still an open question. 
Another interesting possibility is to use AEs to cope with noise by varying manifold dimensionality and by using simple activation functions to introduce sparsity (i.e. ReLU). 

\section{Continual Learning Augmentation}  \label{sec_CLA}
\begin{figure*}
\includegraphics[width=\textwidth]{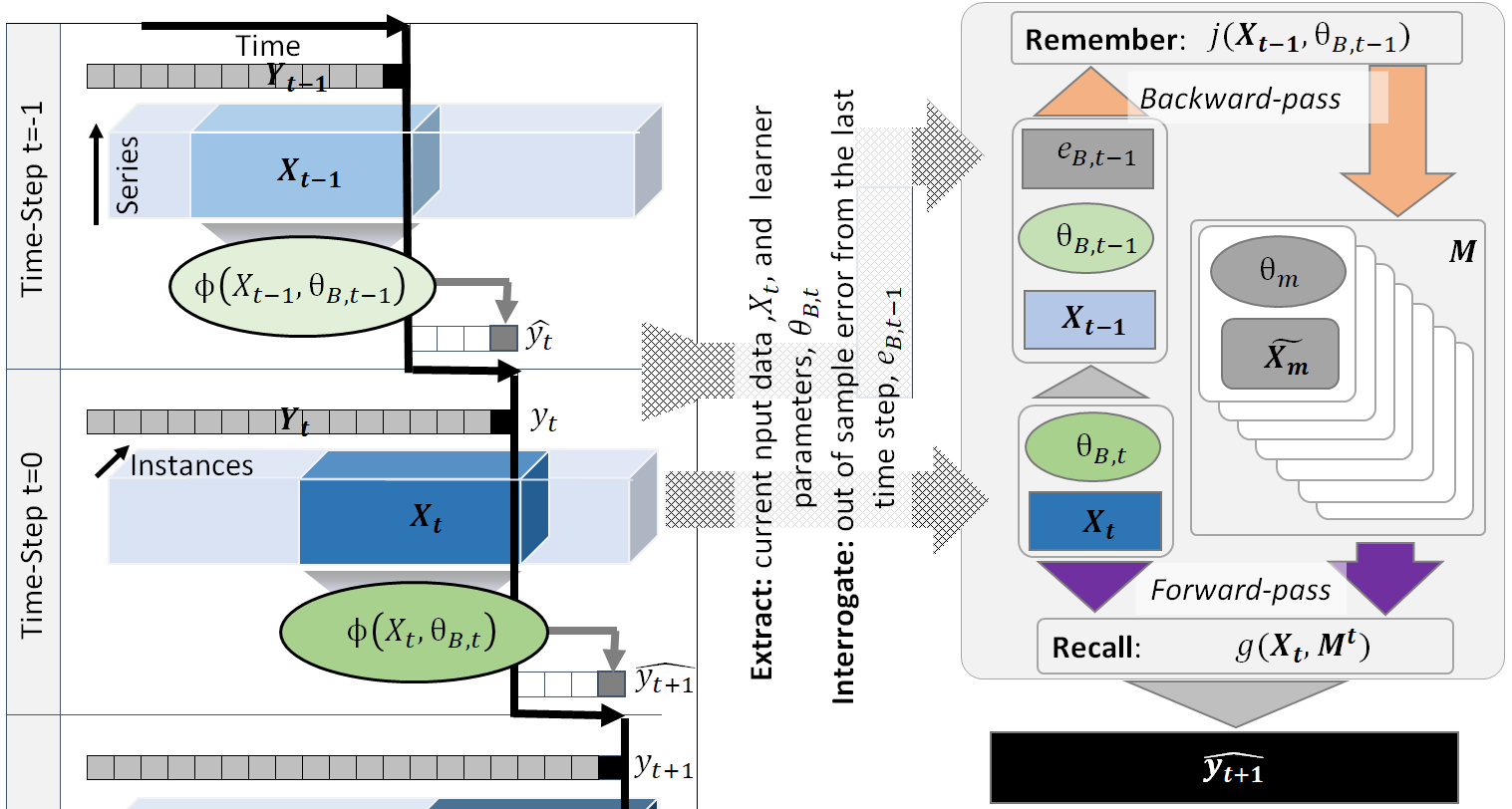}
\caption{
Continual learning augmentation architecture. 
Backward-pass: When $y_{t}$ becomes observable, \emph{remember-gate}, $j$, assesses $|{\epsilon_{B,t-1}}|$ for change and if change has occurred, extracts $\theta_{B,t-1}$ adding a memory column to $\bm{M}$.
Forward-pass: output of learner-memories and base learner are balanced by the \emph{recall-gate}, $g$.
}
\label{CLA_arch2}
\end{figure*}

Continual learning augmentation (CLA)  augments a conventional learner for time-series regression with memory. 
The aim is to allow well understood learners to be used in a CL framework in an interpretable way. 
CLA's memory functions are applied as a sliding window stepping forward through time, over input data of one or more time-series. 
The approach is initialized with an empty memory structure, $\bm{M}$, and a chosen base learner, $\phi$, parameterized by $\theta_B$. 
This base learner can be a sequential approach or a sliding window approach applied to a multivariate input series, $\bm{X}$, with $K$ variables over $T$ time-steps.  
The chosen base learner produces a forecast value $\hat{y}_{t+1}$ in each period as time steps forward. 
A \emph{remember-gate}, $j$, appends a new memory, $\bm{M_t^m}$, to $\bm{M}$, on a \emph{remember cue} defined by the change in the base learner's absolute error at time-point $t$. 
A \emph{recall-gate}, $g$, balances a mixture of current and memory-based forecasts to result in the final outcome, $\hat{y_{t+1}}$.
Figure \ref{CLA_arch2} shows the functional steps of remembering, recalling and balancing learner-memories.

\subsection{Memory management} 
Repeating patterns are required in sub-sequences of the input data to provide memory cues to remember and recall different past states. 
Learner parameters, $\theta_m$, trained in a given past state can then be applied if that state approximately reoccurs in the future. 
When CLA forms a memory, it is stored as a column in an explicit memory structure, similar to \cite{Ciresan_2012}, which changes in size over time as new memories are remembered and old ones forgotten. Each memory column consists of a copy of a past base learner parameterization, $\theta_{m}$, and a representation, $\bm{\widetilde{X_{m}}}$, of the training data used to learn those parameters.
As the sliding window steps into a new time period, CLA recalls one or more learner-memories by comparing the latest input data, $\bm{X_{t}}$, with a representation of the training data stored in each memory column,  $\bm{\widetilde{X_{m}}}$. 
Memories with training data that are more similar to the current input series will have a higher weight applied to their output, $\hat{y}_{m,t+1}$, and therefore make a greater contribution to the final CLA output, $\hat{y}_{t+1}$.

\subsection{Remember-Gate} 

Remembering is triggered by changes in the absolute error series, ${\bm{\epsilon_B}}$, of the base learner as the approach steps forward through time:
\begin{equation}
\bm{\epsilon_B}=\big\{|(\hat{y}_{t}-{y}_{t})|, \dots, |(\hat{y}_{T}-{y}_{T})|\big\}
\end{equation}
CLA interrogates the base learner 
to determine 
changes in out-of-sample error, ${\epsilon_{B,t-1}}$, which are assumed to be associated with changes in state. The remember-gate, $j$, both learns to define and trigger a change which stores a pairing of the parameterization of the base learner, $\theta_{B,t-1}$, and a contextual reference, $\bm{\widetilde{X}_{t-1}}$. 
Figure \ref{CLA_arch2} shows how a change is detected by $j$, which then results in a new memory column being appended to $\bm{M}$: 
\begin{equation}\bm{M} = \big\{(\bm{\widetilde{X}_{1}},\theta_1), \ldots ,(\bm{\widetilde{X}_{M}},\theta_M)\big\}
\end{equation}

Immediately after 
the new memory column has been stored 
a new base learner is trained on the current input, overwriting $\theta_B$. 

Theoretically, for a fair model of a state, $\bm{\epsilon_B}$ would be approximately $i.i.d.$ with a zero valued mean. Therefore the current base model would cease to be a fair representation of the current state when $\bm{\epsilon_B}$ exceeds a certain confidence interval, in turn implying a change in state. 
$J_{Crit}$ is a threshold representing a critical level for $\bm{\epsilon_B}$. 
When the observed absolute error series,$\bm{\epsilon_B}$, spikes above this critical level, it is interpreted as a change in state and also as a remember-cue.

 

%

$J_{Crit}$ is estimated by the hyperparameter $j_{Crit}$. 
This is learned at every time-step to result in a level of sensitivity to remembering that forms an external memory, $\bm{M}$, resulting in the lowest empirical forecasting error for the CLA approach over the study term up until the current time, $t$:

\begin{equation}
J_{Crit} =  
\underset{j_{Crit} \in jgrid} {\operatorname{argmin}}
f(\bm{X_t}, j_{Crit})
\end{equation}
Where $f$ is the CLA approach expressed as a function of the input series, $\bm{X_t}$, and $j_{Crit}$, yielding $\epsilon_{B,t}$ (the absolute error of the base learner at time $t$). $jgrid$ is an equidistant set, between the minimum and the maximum values of $\bm{\epsilon_{B}}$ up until time $t$. This represents a discretization of the empirical distribution of $\bm{\epsilon_{B}}$ from which to empirically solve for $J_{Crit}$. In our testing $jgrid$ was initialised to a value of 20, representing five-percent buckets in the empirical range of $\bm{\epsilon_{B}}$.


\subsection{Recall-Gate} 
The recall of memories takes place in 
the recall-gate, $g$,  
which calculates $\hat{y}_{(t+1)}$,
a mixture of the 
predictions
from the current base learner and from learners stored in memory:  
\begin{equation}
\hat{y}_{(t+1)} = 
g(\bm{X_{t}},\bm{M^t})
\end{equation}
The mixture coefficients are derived by comparing the similarity of the current time varying context $\bm{X_{t}}$ with the contextual references, $\bm{\widetilde{X}_{m}}$, stored with each individual memory. 
Memories that are more similar to the current context have a greater weight in CLA's final outcome. 
\subsection{Recall-gate Similarity Choices} 
Several approaches for calculating contextual similarity are tested separately, using the CLA approach. Each is used to define $\bm{\widetilde{X}_{m}}$, either by simply storing past training examples or by using a process of contextual learning; essentially learning a representation of base learner training data. 
\par
Firstly, ED and then DTW are applied. Both approaches require $\bm{ \widetilde{X}_{m}}$ to be raw training examples, stored in each respective memory column, making both approaches relatively resource hungry. Secondly, AE similarity is used through a process of contextual learning. Rather than needing to store many training examples, only the AE parameters are needed to form a reconstruction of the training data with the disadvantage that an AE must be trained in every time-step, reducing machine-memory usage versus DTW but increasing processing time.
Thirdly we introduce a DTW filtered AE similarity, which is intended to phase adjust the AE reconstruction, we call this \emph{warp-AE}. 
Again, an AE needs to be trained at every time-step but DTW processing expense is less because $N$ in Equation \ref{eq_AEwdist} is equal to the number of securities, while in Equation \ref{eq_DTWdist}, $N$ is generally larger, to provide an effective sample over many security pairings.
We describe each approach in turn.
\par 
ED and DTW are applied to $N$ randomly sampled instances from $\bm{\widetilde{X}_{m}}$ and $\bm{X_{t}}$, sampling over rows (i.e. cross-sectional data), each of which represents a different security at a given point in time:
\begin{equation}\label{eq_EDdist}
\hat{D}_{ED}(\bm{\widetilde{X}_m},\bm{X_t}) = 1/N \sum_
{N}ED(\widetilde{X}_{m,r_1(D)},X_{t,r_2(D)}) 
\end{equation} 

\begin{equation} \label{eq_DTWdist}
\hat{D}_{DTW}(\bm{\widetilde{X}_{m}},\bm{X_t}) = 1/N \sum_
{N}DTW(\widetilde{X}_{m,r_1(D)},X_{t,r_2(D)}) 
\end{equation} 
Where $\hat{D}$ is the dissimilarity, $N$ is the number of samples to take and $r_1(D), r_2(D)$ are random integers between 1 and $D$.

AE similarity is used in a similar fashion to Aljundi et al 2017, \cite{Aljundi17}, using ReLU activations to avoid over-fit. 
However CLA's use of AEs is specifcally designed for state-based CL in noisy, real-world, multivariate time-series. 
To facilitate this application CLA uses additional sparsity loss \cite{OLSHAUSEN19973311_Sparsity} and pure-linear decoder activations to reconstruct the approximate time-series and cross-sectional distributions. 
CLA then uses AE distances to balance memory-recall weightings:

\begin{equation} \label{eq_AEdist}
\hat{D}_{AE}(\bm{\widetilde{X}_{m}},\bm{X_t}) = 1/N \ ED(\bm X_{t}, a(h(\bm X_{t}))) 
\end{equation} 

$ED(\bm X_{t}, a(h(\bm X_{t})))$ is the Euclidean reconstruction loss of the current input, $\bm X_{t}$. $a$ and $h$ are the encoder and decoder functions respectively, $N$ is the number of securities in $\bm{X_t}$. 
\par
\emph{warp-AE} is designed to benefit from AE's lower memory usage while benefiting from the
invariance to time-deformation provided by DTW:

\begin{equation}\label{eq_AEwdist}
\hat{D}_{wAE}(\bm{\widetilde{X}_{m}},\bm{X_t}) = 1/N \ DTW(\bm X_{t}, a(h(\bm X_{t}))) 
\end{equation}  

\par
Each of these four (dis)similarity approaches can be used to determine which memories to recall from $\bm{M}$ and also how to weight the contribution of each to CLA's final outcome, $\hat{y}_{(t+1)}$.
Each similarity approach was tested in turn, in CLA's memory recall-gate, aiming to gain new insights into time-series CL as it is applied to a complex real-world problem.

\subsection{Balancing} 
The base learner and all recalled memories are weighted by similarity to produce CLA's final outcome, using the recall-gate, $g(\bm{X_{t}}, \bm{M})$:  
\begin{equation}
\hat{y}_{t+1}=\sum_{m=1}^{M}\phi(\bm{X_{t}},\theta_{m})\cdot\bigg[1-
\frac{\hat{D}(\bm{\widetilde{X}_{m}},\bm{X_{t}})}{\sum_{m=1}^{M}\hat{D}(\bm{\widetilde{X}_{m}},\bm{X_{t}})}
\bigg]
\end{equation}
Where $M$ is the number of memories in the CLA memory structure, $\bm{M}$.
Previous research indicated balancing was the most powerful approach over selecting the $best$ single memory \cite{Philps_2018}. (Notably, both these balancing approaches significantly outperform equal weighing of all memories, indicating CLA is gaining significantly more than a simple ensemble effect).

\section{Experimental Results} \label{sec_Exp}
\subsection{Setup}

\begin{figure*}[t]
\centering
\includegraphics[width=\textwidth]{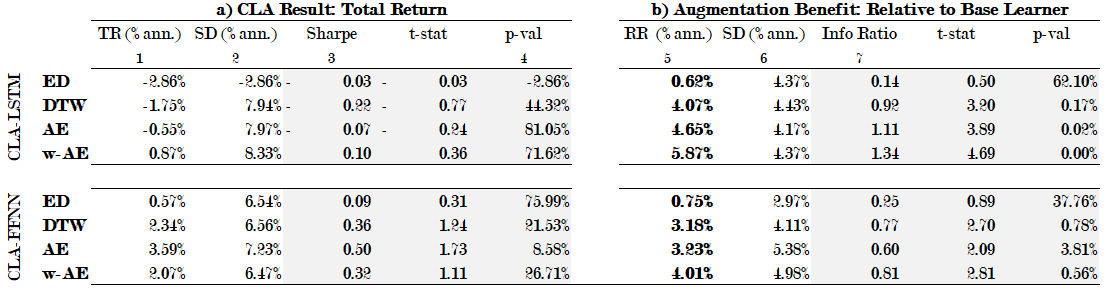}
\caption{Long-short investment simulation results (median over 5 simulation runs). 
Simulations are of long/short emerging market equities, between 2006-2017. No friction costs considered, as analysis of relative (not absolute) performance is the intention.
1) TR: annualized total return, 2) SD: annualized standard deviation, 3) Sharpe ratio: $TR/SD$, 4) p-value of the t-stat for the Sharpe ratio (see below), 5) RR: relative return of CLA versus the base learner performance from same simulation, 6) SD: annualized standard deviation of RR, 7) Information ratio: $RR/SD$.
Results indicate CLA provides an augmentation benefit to FFNN and LSTM. This is concluded from the t-statistics of the Information Ratios, tested to determine a significant difference from zero.
}
\label{CLAChart}
\end{figure*}


CLA is used as a test bed for different learners and similarity approaches in a regression task to forecast future expected returns of individual equity securities. 
This is used to drive equities investment simulations, a real-world task using noisy time-series. 
The data set consists of stock level characteristics at each time-step, for many stocks over many time-steps, as detailed below.
Tests were conducted to show the relative performance of a sliding window base learner, FFNN, and a sequential base learner, LSTM.  Different similarity approaches were also used to drive the memory recall-gate; ED, DTW, AE and \emph{warp-AE}. All combinations of learner and similarity were tested. 
\par
Base learners were batch trained over all stocks at each time-step, forecasting US\$ total returns 12months ahead for each stock. For the sliding window learner a year long, fixed length sliding window of four quarters was used for training and for the sequential learner all historic data up to the current time, was used for training.
A stock level forecast in the top (bottom) decile of the stocks in a time-period was interpreted as a buy (sell) signal. 
\par
Although CLA is designed to use non-traditional driver variables, stock level characteristics are commonly expressed using \emph{factor-loadings}. These were estimated, in-sample at each time-step by regressing style factor excess returns against each stock level US\$ excess return stream over a 36month window: 
$r_{i,t} = \alpha_{i,t} + \beta_{MKT,i,t}x_{MKT,t} + \beta_{VAL,i,t} x_{VAL,i,t} + \epsilon_{i,t}$,
where $r_{i,t}$ is the excess return of stock $i$ in period $t$, $x_{MKT,t}$ is the excess return of the Emerging Market Equities Index, $x_{VAL,t}$ is the relative return of the Emerging Market Value Equities Index. 
\par
Stock level factor loadings populated a matrix, $\bm{X}$, which comprised the input data. 
Each row represented a stock appearing in the index at time $t$ (up to 5,500 stocks) and each column related to a coefficient calculated on a specific time lag. 
$\bm{X}$ resulted from a fifth and ninety-fifth percentile winsorizing of the raw input to eliminate outliers. 
\par
Long/short model portfolios were constructed (i.e. rebalanced) every six months over the study term, using equal weighted long positions (buys) and shorts (sells). 
The simulation encompassed 5,500 equities in total, covering 26 countries across emerging markets, corresponding to an Emerging Market Equities Index between 2006-2017.
To account for the sampling approach used for ED and DTW similarities and differences in random initialisation of neural components, several simulations were carried out per test.


\subsection{Simulation Results}

CLA results showed a significant augmentation benefit for both base learners (see Figure \ref{CLAChart} b), while all similarity approaches performed better than ED. Unaugmented base learners showed mixed results.

\par Sliding window learner tests, CLA-FFNN, outperformed all the equivalent sequential learner tests, CLA-LSTM, in terms of total return (TR) while Sharpe ratios (see Figure \ref{CLAChart} a) were superior also (although no positive Sharpe ratio was significant at the 5\% level). However, augmentation benefit, gauged by relative return (RR) and information ratio (Info Ratio), was superior for CLA-LSTMs (Figure \ref{CLAChart} b), with most augmentation tests for both learners statistically significant at the 5\% level. 
In these tests, although CLA-LSTM saw a better augmentation benefit (RR), CLA-FFNN saw the strongest outright performance (TR), followed by unaugmented FFNN (given by TR-RR), then CLA-LSTM. By far the weakest outright performer (TR) was unaugmented LSTM (given by TR-RR).

\par Tests of different similarity approaches, used in the recall-gate, saw ED under-perform DTW in TR terms and also in terms of augmentation benefit. This was true for both learners tested. This would imply that the invariance to time-deformation DTW provides, is an important consideration in a real-world context. AE similarity tests showed higher TRs than DTW and demonstrated statistically significant augmentation benefits at the 5\% level for both learners, indicating that AE similarity is an appropriate approach to use in this context. \emph{warp-AE} generated the highest RR and information ratios of all similarity tests, implying that adding a DTW filter to AE similarity was the most interesting similarity approach tested.

\subsection{Interpretable Memory}
\begin{figure*}
\begin{center}
\textbf{Interpretable Memory}\par\medskip
\includegraphics[width=0.9\textwidth]{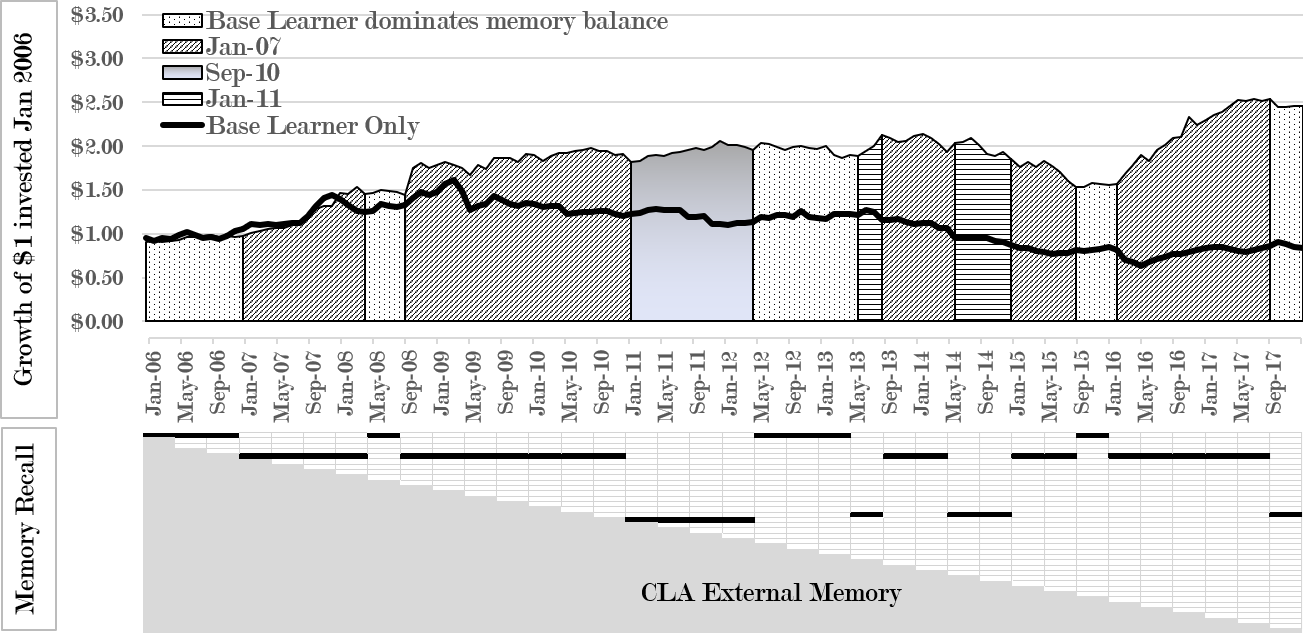}
\caption{Interpretable memory.  Top: Investment returns of an unaugmented FFNN base learner (line) shown next to CLA-FFNN returns (area), from the same CL instance,  separated by memory contribution. Lower chart: representation of the CLA memory structure, where each row in the expanding memory-triangle represents a step-forward and therefore a potential remember-cue. Highest weighted recalled memories are show as horizontal bold lines. 
}
\label{CLAmem}
\end{center}
\end{figure*}

CLA produces outcomes that can be attributed to specific memories. Figure \ref{CLAmem} shows an example of one of the simulation runs, CLA-FFNN with AE similarity. The upper panel shows hypothetical investment returns of the unaugmented base learner (line chart) compared to those of the CLA augmented base learner (area). 
The lower panel shows CLA's expanding-memory-triangle, expanding by one row at each time-step forwards, indicating the possibility of the system adding a new memory at each step. Note the black horizontal lines, showing when certain memories were recalled and applied. In this example, at least three memories are remembered and recalled at later times. Qualitatively, a memory remembered in January 2007, a period of turbulence in financial markets, adds materially to the strategy return in later periods, when recalled. It proves more appropriate than the base learner in the period of the 2008 financial crisis and its aftermath involving concerted fiscal stimulus (Sept 2008-Dec 2010). It was again recalled in 2013 and in 2015/2016, periods where fiscal stimulus dominated market returns (in Europe and China respectively). The sparsity of the memory structure is also notable, with only three principle memories recalled/used. 
As a rule of thumb, fewer memories will be remembered/recalled for noisier data-sets. This occurs through either a lower learned sensitivity to remembering, from a higher learned value of $j_{Crit}$, or through less recalling, from less discernible contexts. 

\section{Conclusion} \label{sec_Conclusion}
We have empirically demonstrated that when applied to a real-world financial task involving noisy time-series with a large cross-sectonal component, a CL augmented sliding window learner (CLA-FFNN) is superior to both an LSTM learner and to a CL augmented LSTM learner (CLA-LSTM).

Testing of different similarity approaches, applied to a recall-gate, showed poor performance of simple Euclidean distance (ED) when compared to dynamic time warping (DTW). 
This strongly implies that the timing of data-points is important in this task and likely in other real-world problems involving noisy time-series. Simulation tests also showed that AE similarity is a good alternative to DTW. 
\emph{warp-AE} was proposed to benefit from both DTW's time-deformation invariance and AE's lower machine-memory requirements, an approach that produced the strongest augmentation benefit. 
We also show that time-series CL not only outperforms LSTM and FFNN base learners but can provide a transparent explanation for which memory did what and when. 
\par
In summary, the most successful CL augmentation choice was found to be a sliding window CLA-FFNN learner combined with a recall-gate using \emph{warp-AE} similarity. These tests also affirm Continual Learning Augmentation (CLA) as a real-world time-series CL approach, with the flexibility to augment different types of learners. 

\section{Bibliography}
\printbibliography[heading=none]

\end{document}